# TBX goes TEI

Laurent Romary

## Abstract

This paper presents an attempt to customise the TEI (Text Encoding Initiative) guidelines in order to offer the possibility to incorporate TBX (TermBase eXchange) based terminological entries within any kind of TEI documents. After presenting the general historical, conceptual and technical contexts, we describe the various design choices we had to take while creating this customisation, which in turn have led to make various changes in the actual TBX serialisation. Keeping in mind the objective to provide the TEI guidelines with, again, an onomasiological model, we try to identify the best comprise in maintaining both the isomorphism with the existing TBX Basic standard and the characteristics of the TEI framework.

## Keywords

## Why adapting TBX for the TEI guidelines?

### A brief history of the TBX lineage

TBX (TermBase eXchange) is the name of a representation format for terminological data that has been published, in its full form, as an ISO standard in 2008. It is the result of more than twenty years of standardisation activities in the domain of computerized terminologies that have taken place in ISO, the TEI[1] consortium and the LISA[2] association. It is difficult to understand both the importance of this standard and the necessity to provide a simplified version integrated in the TEI guidelines without keeping in mind the overall history that lead to its publication. The major steps in such a history can be outlined as follows:

- ISO 6156:1987 (Mater), which describes a format for representing terminological information on magnetic tapes, is probably the first international standard ever that dealt with the interchange of lexical data. Designed as a flat field-based representation in a pre-SGML world, the format is further developed to be applicable on microcomputers (MicroMater; see Melby, 1991);

- The experience gained with the work on Mater and MicroMater lead Alan Melby and several other colleagues[3] to take part to the Text Encoding Initiative and put together a specific chapter[4] of the TEI guidelines dedicated to the representation of terminological data. This first attempt to have an SGML-based representation integrated in the TEI framework remained there until the P4 edition when the TEI guidelines switched to an XML conformant

---

representation and it was observed how updated the format was with regards more recent developments. Still, As can be seen from Figure 1, the TEI representation, despite its missing language section, contains most of the elements that one can find in the current TBX Basic format;

- ISO 12200 (Martif), published in 1999, resulted from a wish to take up and improve the TEI proposal, but also to make is an independent initiative[5] under the auspices of ISO. Beyond the introduction of a language section, Martif also embeds terminological entries within a document structure strongly inspired from the TEI one (e.g. the header-text organisation; entries embedded within a <text> and <body> hierarchy);

- In 1999, ISO also published an important standard, ISO 12620:1999[6], which compiled a set of reference descriptors (or *data categories*) to be used in any kind of terminological database. This standard represented the first abstraction over specific terminological formats as well as a tool for specification and customisation in the domain of termbank design. All descriptor we will use in this paper and in the TEI customisation we describe originate in this standard;

- ISO 16642 (TMF), published in 2003, brought another level of abstraction by describing a meta-model and modelling mechanisms independently of any specific XML format. Interoperability across terminological formats is basically ensured through compliance to ISO 16642 and ISO 12620:1999;

- In the period that followed, further work was carried out within LISA to define an optimal follower to Martif, which lead to the publication of the first version of TBX (TermBase eXchange) in 2007, which in turn was published as ISO standard 30042 in 2008.

As published in 2008, the TBX standard appears to be a comprehensive integration of the progress made over the years in understanding the variety of needs in the domain of digital terminology management. Still, the cumulative design that we have tried to outline before made it a complex object with two many options (e.g. two elements <tig> and <ntig> for implementing the term section component) and above all a proprietary technological implementation (reflected in the so-called XCS parameter files) for customising the necessary subsets needed specific implementations. This situation has slowed down the actual take-up of the format dramatically, with the emergence of other initiatives such as SKOS within the W3C that could completely ignore the benefits a proper terminological representation of the corresponding data. In parallel, the quick uptake of digital methods in the humanities, where several research communities such as field linguists, epigraphists or classicists (among others) have a need to record onomasiological data associated to their primary sources, made it necessary to consider providing them back with an accessible chapter in the TEI guidelines.

```
<termEntry>
      <admin type='domain'> appearance of materials </admin>
      <tig lang=en>
            <term> opacity </term>
            <gram type=pos> n </gram>
            <descrip type='definition'> degree of obstruction to the
            transmission of visible light </descrip>
            <ptr type='bibliographic' target='ASTM.E284'>
            <admin type='responsibility' resp='ASTM E12'> </admin>
```

```
        </tig>
        <tig lang=de>
            <term> Opazität </term>
            <gram type=pos> n </gram>
            <gram type=gen> f </gram>
            <descrip type='definition'> Maß für die
            Lichtdurchsichtigkeit </descrip>
            <ref type='bibliographic' target='HFdn1983'> p. 383 </ref>
            <admin type='responsibility' resp='DIN TC for paper
            products'></admin>
        </tig>
        <tig lang=fr>
            <term> opacité </term>
            <gram type=pos> n </gram>
            <gram type=gen> f </gram>
            <descrip type='definition'> rapport du flux lumineux
            incident au flux lumineux transmis ou réfléchi
            par un noircissement photographique </descrip>
            <ptr type='bibliographic' target='HJdi1986'>
            <admin type='responsibility' resp='C.I.R.A.D.'> </admin>
        </tig>
</termEntry>
```

**Figure 1: A TEI P3 example of terminology encoding (source: http://www.tei-c.org/Vault/GL/P3/TE.htm)**

## The TEI perspective

The text Encoding Initiative was initiated in 1987, when a group of experts in charge of several major literary text archives met to establish the foundation of what was about to become the most ambitious standardisation activity in the humanities. Under the auspices of the two major learned societies of that time, the ACH[7] and the ALLC[8], it produced reference guidelines that, since their first stable edition in 1992 (see Burnard & Sperberg-McQueen, 1995), evolved to incorporate both the most recent technological evolutions and the continuous feedback from its constantly growing user community.

There are several reasons why the paths of both terminology standards and the TEI crossed at some point:

- The Text Encoding Initiative had created a dynamics combining both a strong political (and financial) support as well as a high level of technological awareness. It brought together experts from many different fields ranging from poetics to computational linguistics, with also an informed contribution from the library community;
- The first important decision related to the design of the TEI guidelines was to adopt the recently published ISO SGML standard (ISO 8879, published in 1986!) which offered, despite its complexity, a then unique platform for specifying and customising complex document structures;
- From a content point of view, the TEI was organised into working groups dedicated to various domains of text technology and representation. For lexical data, there was an ambition to cover both semasiological and onomasiological representation models with on the one hand a group dedicated to "print dictionaries" and, on the other hand, one working on a future chapter "terminologies".

In this somehow favourable double context, the present paper describes what we think is an optimal strategy to both enrich the TEI guidelines with a proper means of representing onomasiological (terminological) data and contribute to the wider

---

dissemination of best practices in this domain as reflected by the existing portfolio of international standards.

### Filius prodigus

The main idea is to take of the simplified representation of a terminological entry as depicted in the TBX Basic proposal and incorporate in at any place in a TEI document where a semasiological representation (mainly embodied in the TEI guidelines by the <entry> element) could also take place. The choice of TBX Basic, a very small subset of TBX originally developed by LISA, is justified by the need to achieve the best compromise between simplicity and conformance to the ISO standard. It is indeed essential, when considering this proposal, to keep in mind that the target user group is that of scholars with enough digital awareness to use the TEI guidelines, but with little or no background in terminology management. The resulting format should thus be both easy to use and compatible with the TEI cultural background.

The choice thus made can be seen as a start for further iteration between the TEI and TBX since it is based on two approximate assumptions. First, it considers that nothing should be kept from the existing TBX macro-structure and that the TEI architecture takes priority. TBX is somehow forced to fit back into the TEI guidelines. Second, by taking up TBX basic, no preliminary analysis is made as to the type semasiological use cases that would be adapted to the TEI user community. In particular, more precise insights should be gained to identify which data categories are actually needed.

## A basic overview of the available building blocks

### The TEI document architecture

The TEI guidelines contain more than 500 elements to mark-up textual documents at various level of structural granularity and to take into account many different textual genres such as manuscripts, dramas, speech transcriptions or dictionaries. Its main document structure, depicted in Figure 2, combines a mandatory header (<teiHeader> element), grouping together all the meta-data attached to the document, with the actual content (<text>). This content can be further decomposed as a <front>, a <body> and a <back> element, in reference to the traditional book structure.

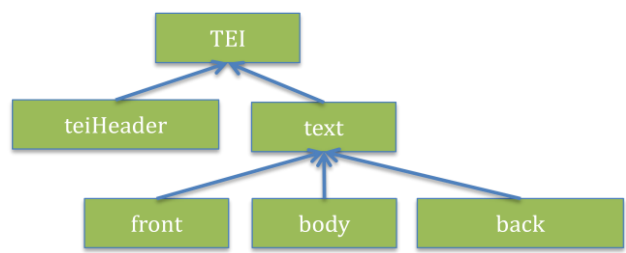

**Figure 2: TEI document architecture**

The TEI vocabulary provides a variety of means to encode a document, which we can sketch according to the following main categories:

- Description of the structure of a text by means of the generic <div> element;
- Organisation of the content along paragraph level objects such as lists, figures, tables, etc.
- Inline annotation elements to mark-up specific linguistic segments (highlighted object, foreign expressions) or reference to entities (names, dates, numbers);

- Domain specific constructs for dealing with turns in speech transcription, dictionary entries, etc.
- General-purpose representation objects such as bibliographical descriptions.

All these elements are part of a reference framework from which a given project should select which elements are needed for its purpose. Indeed the TEI guidelines provide specific mechanisms to express such customisations as described in the next section.

## The TEI specification framework

The TEI guidelines can be seen from two different angles. First, as the basis of an XML representation format, they provide the technical constraints to control the validity of TEI conformant document instances. Second, they are delivered with an extensive prose description that informs users about the logic of the guidelines as well as the most appropriate way(s) to use them to represent specific textual phenomena[9]. Still, these two views are not split into two separated objects but indeed integrated within one single specification, from which one view and the other can be automatically generated. This mechanism, in line with the concept of literate programming (Knuth, 1992), relies in the existence of an underlying specification language named ODD (One Document Does it all), which is itself expressed in TEI.

In the TEI infrastructure, each element is thus defined as an ODD specification providing all the necessary information both to control its (XML) syntactic behaviour and to generate the corresponding documentation. Such information comprises a gloss, a definition, the technical description of its content model, the various attributes it can bear and one or several example of its usage.

Besides, the TEI framework offers two additional mechanisms that are central in providing the global coherence of the TEI guidelines: Classes and Modules.

There are two types of classes: attribute classes, which groups together attributes used within the same elements[10] and model classes, which group together elements that a related semantics and occur at the same places in a document. The latter are means to simplify the expression of content models, but also to facilitate the customisation process by simply adding or removing an element from a class.

Modules are more global objects intended to group together coherent sets of elements designed for a similar purpose. Typically all the specific elements for dictionary encoding are grouped together in one single class.

These various mechanisms form the basis of the customisation mechanisms in the TEI guidelines based upon the ODD language, namely by selecting a group of modules for a given representation objective and within each module, keeping or discarding elements in the content model by editing the appropriate model classes.

## The TBX entry model

The TBX entry model is organised in strict compliance to ISO 16642 (TMF) such that the three core components of TMF, namely *Terminological Entry*, *Language Section* and *Term Section* are implemented respectively by means of the three elements: <termEntry>, <langSet> and <tig>[11].

---

[9] The TEI guidelines contain in particular a wealth of examples for each element and the major constructs they allow.

[10] For instance, the class att.global, which contains general purpose attributes such as the W3C @xml:id, @xml:lang or the TEI generic @n (local numbering), @rend (rendering information).

The descriptive objects associated to these three levels are made of a) specific elements such as <term>, <note>, <ref> and <xref> and b) so called meta data-elements that may express a wide range of possible data categories, namely <admin>, <descrip> and <termNote>. For instance, <descrip type="definition"> is the TBX construct to represent a definition at the level of <termEntry> or <langSet>.

Figure 3 shows a typical (reduced) example of a TBX entry, where we can see how specific descriptions can also be embedded within grouping elements (<descripGrp>), when they need to be refined by additional information.

```
<termEntry xmlns="http://www.tbx.org">
    <descrip type="subjectField" xml:lang="fr">Industrie mécanique</descrip>
    <langSet xml:lang="de">
        <descripGrp>
            <descrip type="definition">endloser Riemen mit trapezförmigem
Querschnitt, der auf zwei Riemenscheiben mit Eindrehungen läuft</descrip>
            <admin type="source">De Coster, Wörterbuch,
Kraftfahrzeugtechnik, SAUR, München, 1982</admin>
        </descripGrp>
        <note>wird zum Antrieb der Lichtmaschine, des Ventilators und der
Wasserpumpe benutzt</tei:note>
        <tig>
            <term>Keilriemen</tei:term>
            <admin type="source">De Coster, …</admin>
        </tig>
    </langSet>
    <langSet xml:lang="fr">
        <descripGrp>
            <descrip type="definition">courroie sans fin …</descrip>
            <admin type="source">De Coster, …</admin>
        </descripGrp>
        <tig>
            <term>Keilriemen</tei:term>
            …
        </tig>
    </langSet>
</termEntry>
```
Figure 3: Example of a TBX Basic entry (source: http://iate.europa.eu)

## Technical integration

### TBX in ODD

In order to integrate the TBX entry construct in the TEI framework, we have actually created a full ODD specification for it, which facilitates and documents the merging process. This ODD specification of TBX, which is delivered with the pre-print of this publication[12], is based on a simple architecture depicted in Figure 4 and strongly inspired from the TBX original organisation. Each level of the TBX terminological entry model is specified as a combination of members from the class *model.auxInfo* (whose member are, by default, <admin>, <descrip>, <descripGrp>, <transacGrp>, <note>, <ref>) and element(s) required at the child level.

For instance, the content model of <tig> is simply described as:

```
<rng:group>
    <rng:ref name="term"/>
    <rng:zeroOrMore>
        <rng:ref name="termNote"/>
    </rng:zeroOrMore>
    <rng:zeroOrMore>
        <rng:ref name="model.auxInfo"/>
```

---

[11] All TBX elements are implemented within the TBX namespace: "http://www.tbx.org"
[12] http://hal.inria.fr/hal-00950862

```
        </rng:zeroOrMore>
    </rng:group>
```

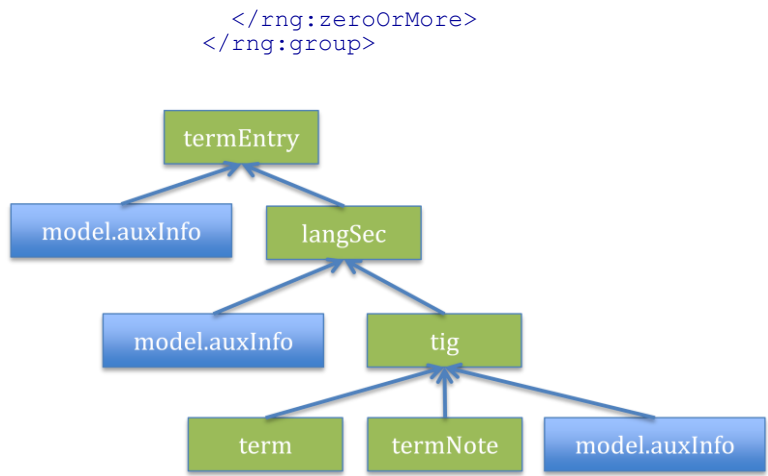

**Figure 4: Overall architecture of the ODDD model of a TBX entry**

## Modernizing and complementing the TBX technical setting

This first level of specification of the TBX entry in ODD has some immediate impact on its actual XML properties. First, all elements in this specification, like any element from the TEI guidelines are made member of the att.global attribute class to supply them with so-call global attributes, in particular the generic W3C attributes @xml:id, @xml:lang, @xml:base, @xml:space. Second, all elements originally bearing a @target attribute in TBX are made members of att.pointing, the corresponding attribute class where this attribute is defined. The consequence here is that the old-fashioned ID/IDREF pointing mechanism in TBX is refreshed into a general URI based reference system. This change also makes the difference between internal (<ref>) and external (<xrev>) reference obsolete leading to the disappearance of <xref>.

## Conflicting elements and attributes

Several elements[13] and attributes[14] have the same name as other elements and attributes in the TEI guidelines. This could be seen as a non-issue when dealing with these in their appropriate namespaces, thus avoiding any syntactic conflict. Still, we have considered it important to provide a vision for making such objects in TBX and in the TEI converge, and this for the following reasons:

- For historical reasons, the TEI has been a strong inspiration for TBX, and the various TBX instances of the duplicate elements are indeed clones of the corresponding TEI ones, with a very similar underlying semantics;

- From a user point of view, it would be difficult to systematically keep both of these elements and attributes in the same XML document model, without spending a lot of energy trying to elicit the actual differences;

- The TBX instances are often simplified versions of their TEI counterparts, both from the point of view of their content models and the attributes they may bear;

- Even more, some of the TBX elements or attributes have departed from their original semantics (<hi>) or lost base with the XML technological evolutions (<ref> or @target, still relying on the deprecated ID/IDREF mechanisms);

---

[13] term, ref, hi, foreign
[14] type, target

- Finally, the perspective of merging part of the TBX vocabulary in the TEI framework in the long run, should lead us to anticipate a necessary convergence.

These various arguments have lead us to replace all TBX instance of the duplicated by their TEI counterparts, even if we are fully aware of the drawback resulting from the frequent alternation of namespaces within a terminological entry.

## Enriching element contents

One important aspect of the TEI infrastructure is that it provides a large range of elements for the inline annotation of textual content. Such elements are indeed essential for the precise mark-up of all sorts of entities mentioned in a text (persons, places, etc.), linguistic phenomena (mentioned references, foreign expressions) but also technical means like inline cross-referencing. In this respect, the TBX framework is relatively poor since it only allows the following elements in the content model of plain text components[15]:

- A group of so-called *meta-markup* elements (<bpt>, <ept>, <ph>) designed to encapsulate non XML mark-up from an primary text source. These elements have also become quite standard practice in localisation or translation memory documents;
- Two specific inline elements: <hi> for highlighting segments of texts, optionally cross-referencing an external object (with a @target attribute) and <foreign> to annotate foreign words or expressions inline.

The process of creating a customisation incorporating TBX entries in TEI documents offers the perspective of enlarging the number and types of such components while preserving backward compatibility with the legacy model of such textual elements. The proposed customisation implements this extension in two steps:

- The creation of a class *model.metaMarkup* grouping together all above-mentioned meta-markup elements;
- Express the content model of textual elements as a combination of text, members of the *model.metaMarkup* class and the existing *model.limitedPhrase* from the TEI infrastructure.

```
<rng:zeroOrMore>
  <rng:choice>
    <rng:text/>
    <rng:ref name="model.limitedPhrase"/>
    <rng:ref name="model.metaMarkup"/>
  </rng:choice>
</rng:zeroOrMore>
```

By doing so the content model of text elements changed in several ways:

- The meta-markup elements are available in the TBX namespace;
- Textual elements are provided with a wide variety of annotation objects;
- The TBX <hi> and <foreign> are replaced by their TEI counterparts. This removes the @target attribute from <hi>, which is somehow fortunate since it guides users towards adopting more appropriate elements (such as <ref>) for cross-referencing.

---

[15] <admin>, <descrip>, <note>, <ref> and <termNote>; <term> has a specific status as it only allows <hi> in its content.

## How can TBX benefit from going TEI

Beyond the obvious gain in visibility that the TBX standard can benefit from by being linked to the widely adopted TEI framework, we would like to focus on the various ways the work described in this paper could positively impact on the future technical definition of TBX.

The main proposal for TBX in managing its evolution within a more sustainable framework would be to re-align its document macro-structure to that of the TEI. Indeed, the whole document model down to <termEntry> in TBX is strongly inspired from the TEI guidelines, but suffers from both its oversimplification and from the fact that it cannot automatically benefit from existing components, and further improvements, in the TEI guidelines that would immediately be valuable to TBX (e.g. software description, bibliographical sources, etc.). By re-using the TEI model in its reference namespace, with the appropriate customisation in order to avoid cluttering the header with constructs that have nothing to do with terminology management, TBX would benefit from a rich and maintained meta-data environment while keeping the freedom of defining the standard as a specialized subset.

One real challenge for the experiment we are carrying out here is to find a strategy for the various domains where the TEI provides alternative, and essentially more precise, constructs than TBX for the same types of objects.

A first example are the grammatical features in the TEI guidelines, which offer a wide variety of elements for providing part of speech, gender, number, etc. information, possibly grouped together within a <gramGrp> container. For users accustomed to the TEI background, such descriptors would be obvious alternatives to the use of the <termNote> element from TBX. Should we provide both possibilities in the long term? Should we suggest the use of <gramGrp> in TBX? Or should we just consider that the use of one or the other corresponds to two dialects (TEI-TBX vs. mainstream TBX) with explicit mappings between the two.

In the same way, the bibliographical descriptions in TBX are very shallow and only allow plain text content such as:

```
<admin type="source">De Coster, Wörterbuch, Kraftfahrzeugtechnik, SAUR,
München, 1982</admin>
```

whereas the TEI could possibly offer any level of representation for such object by means of unstructured (<bibl>) or structured (<biblStruct>) representation, for instance:

```
<bibl><author>De           Coster</author>,            <title>Wörterbuch,
Kraftfahrzeugtechnik</title>,              <publisher>SAUR</publisher>,
<pubPlace>München</pubPlace>, <date>1982</date></bibl>
```

There are several similar cases that we could point out, in particular with regards the descriptions of the forms associated to a term (pronunciation, transliterations, inflexions). The precise analysis of all the potential conflicts goes beyond the scope of this version of the paper, but we consider that the answer should not be made issue per issue, but be based upon global scenarios as to what is the optimal implementation in the TEI framework, what level of interoperability we want to keep with mainstream TBX and what the TBX standard itself could actually incorporate in the future.

## Where should we go from here?

We have described in this paper an attempt to create a missing component in the TEI guidelines that would provide an onomasiological representation for lexical data in complement to the existing "dictionaries" chapter. This component is strongly based upon the TBX standard, but we have made the choice of defining of a specific blend of TBX rather than just plugging-in TBX entries within a TEI document. This is, to our view, an essential step if we want acceptance of such representations within the largest TEI user community covering all fields in the humanities. Still, in conformance to the principles of ISO 16642 (TMF) we have tried to ensure maximal isomorphism between the blend we have defined and the original TBX standard, while identifying the situations where the TEI could bring even more encoding precision.

The vision that needs to be developed in the long term for the TEI guidelines should provide a clear guidance as to which model can be used for which purpose in a Digital Humanities scenario. In this context, it is important to remember that TBX Basic was conceived for a specific class of applications in terminology management and not to offer a simple framework for onomasiological representations[16]. Further work should also contribute to have a better understanding on the optimal articulation between semasio-lexical representations and onomasio-terminogical ones in the humanities. Such work could be inspired by seminal works such as (Melby & Wright, 1999) or the ongoing Ontolex initiative within W3C.

---

[16] We have not gone in the details of the TBX basic data categories, but we can observe how such categories as *projetSubset*, with its industrial background, probably out of scope of a humanities application.

## Acknowledgements

This work has benefited from the support og the French ANR TermITH project (http://www.atilf.fr/ressources/termith/).